\documentclass[conference]{IEEEtran}
\IEEEoverridecommandlockouts
\usepackage{enumerate}
\usepackage{cite}
\usepackage{amsmath,amssymb,amsfonts}
\usepackage{algorithmic}
\usepackage{graphicx}
\usepackage{textcomp}
\usepackage{xcolor}
\usepackage{balance}

\usepackage{caption}
\usepackage{subcaption}

\usepackage{tikz}
\usetikzlibrary{svg.path}
\tikzset{
  orcidlogo/.pic={
    \fill[orcidlogocol] svg{M256,128c0,70.7-57.3,128-128,128C57.3,256,0,198.7,0,128C0,57.3,57.3,0,128,0C198.7,0,256,57.3,256,128z};
    \fill[white] svg{M86.3,186.2H70.9V79.1h15.4v48.4V186.2z} svg{M108.9,79.1h41.6c39.6,0,57,28.3,57,53.6c0,27.5-21.5,53.6-56.8,53.6h-41.8V79.1z M124.3,172.4h24.5c34.9,0,42.9-26.5,42.9-39.7c0-21.5-13.7-39.7-43.7-39.7h-23.7V172.4z} svg{M88.7,56.8c0,5.5-4.5,10.1-10.1,10.1c-5.6,0-10.1-4.6-10.1-10.1c0-5.6,4.5-10.1,10.1-10.1C84.2,46.7,88.7,51.3,88.7,56.8z};
  }
}
\newcommand\orcidicon[1]{\href{https://orcid.org/#1}{\mbox{\scalerel*{
\begin{tikzpicture}[yscale=-1,transform shape]
\pic{orcidlogo};
\end{tikzpicture}
}{|}}}}
\newcommand\copyrighttext{
\footnotesize \textcopyright 2021 IEEE. Personal use of this material is permitted. Permission from IEEE must be obtained for all other uses, in any current or future media, including reprinting/republishing this material for advertising or promotional purposes, creating new collective works, for resale or redistribution to servers or lists, or reuse of any copyrighted component of this work in other works.}
\newcommand\copyrightnotice{\begin{tikzpicture}[remember picture,overlay]
\node[anchor=south,yshift=10pt] at (current page.south) {\fbox{\parbox{\dimexpr\textwidth-\fboxsep-\fboxrule\relax}{\copyrighttext}}};
\end{tikzpicture}}

\begin{document}
\bstctlcite{IEEEexample:BSTcontrol}

\title{
A Safe Reinforcement Learning Architecture for Antenna Tilt Optimisation
}

\author{
\IEEEauthorblockN{
    Erik Aumayr\IEEEauthorrefmark{1},
    Saman Feghhi\IEEEauthorrefmark{1},
    Filippo Vannella\IEEEauthorrefmark{2}\IEEEauthorrefmark{3},
    Ezeddin Al Hakim\IEEEauthorrefmark{2},
    Grigorios Iakovidis\IEEEauthorrefmark{2}\IEEEauthorrefmark{3}}
\IEEEauthorblockA{
\IEEEauthorrefmark{1}Network Management Research Lab, LM Ericsson, Athlone, Ireland\\
\IEEEauthorrefmark{2}Ericsson Research, Stockholm, Sweden\\
\IEEEauthorrefmark{3}KTH Royal Institute of Technology, Stockholm, Sweden\\
\\
}
}

\maketitle
\copyrightnotice
\footnotetext[4]{This work was partially supported by the Wallenberg AI, Autonomous Systems and Software Program (WASP) funded by the Knut and Alice Wallenberg Foundation.}

\begin{abstract}
Safe interaction with the environment is one of the most challenging aspects of Reinforcement Learning (RL) when applied to real-world problems.
This is particularly important when unsafe actions have a high or irreversible negative impact on the environment.
In the context of network management operations, Remote Electrical Tilt (RET) optimisation is a safety-critical application in which exploratory modifications of antenna tilt angles of base stations can cause significant performance degradation in the network.
In this paper, we propose a modular Safe Reinforcement Learning (SRL) architecture which is then used to address the RET optimisation in cellular networks. In this approach, a safety shield continuously benchmarks the performance of RL agents against safe baselines, and determines safe antenna tilt updates to be performed on the network.
Our results demonstrate improved performance of the SRL agent over the baseline while ensuring the safety of the performed actions.
\end{abstract}

\section{Introduction}
\label{sec:intro}

The scale of modern cellular networks and user service demands make performance optimisation more challenging than ever.
Network configuration must be adjusted automatically and in (near) real time to ensure a high level of Quality of Service (QoS) to each User Equipment (UE).
The Remote Electrical Tilt (RET) angle of antennas, defined as the change in the vertical orientation of antennas by an electrical design, is one of the most important variables to control for Self-Organising Networks (SONs). The goal is to find an optimal trade-off for some of the network Key Performance Indicators (KPIs), such as \textit{quality}, \textit{coverage} and \textit{capacity} \cite{Buenostado17}.
For example, an increase in antenna downtilt correlates with a stronger signal in a more concentrated area as well as higher capacity and reduced interference radiation towards other cells in the network.
However, excessive downtilting can result in insufficient coverage in a given area, with some UE unable to receive a minimum Reference Signal Received Power (RSRP).



Many existing solutions to downtilt adjustment use rule-based algorithms to optimise the tilt angle based on the historical network performance.
These rules are usually created by domain experts, and thus lack the scalability and adaptability required for modern cellular networks.
Developments in Reinforcement Learning (RL) \cite{kaelbling1996reinforcement} introduce novel data-driven solutions where RET optimisation is performed by an agent that continuously performs tilt angle adjustments and observes the changes in the environment in order to learn the optimal action for each network state \cite{razavi2010fuzzy, fan2014self, guo2013spectral, balevi2019online}.

However, when interacting with a sensitive environment, such as a cellular network, even small unsafe modifications can lead to drastic performance loss and disruption, which makes traditional RL solutions risky and impractical.
Therefore, RL has found limited application in mission-critical use cases.
These limitations lay grounds for the study of Safe Reinforcement Learning (SRL), where the RL agent is restricted in its exploration of the environment. 
The restriction can be achieved by various safety mechanisms that minimise the adverse effects of unsafe actions \cite{garcia2015comprehensive}.
Our approach relies on existing baseline strategies that are known to be \textit{safe} with respect to a minimum level of performance, such as rules created by a domain expert.
When an action that is proposed by the RL agent performs worse than the baseline, it is deemed \textit{unsafe}, and the baseline action may be executed instead.
Hence, we define \textit{safety} as performing at least as well as a baseline whose performance is known to be acceptable.

In this paper, we propose a comprehensive modular architecture for SRL and demonstrate its viability on the RET optimisation use case.
The architecture consists of a safety shield, safety logic modules, a set of known safe baselines, and one or more RL agents.
The \textit{safety shield} wraps around the target environment with the goal of decoupling the RL agent from the environment, thereby preventing the RL agent from directly and uncontrollably affecting the environment.
The \textit{shield logic modules} provide methods to evaluate actions suggested by RL agents and baselines, and to choose which action should be performed on the environment.
The \textit{baselines} are considered safe on at least a subset of the state-action space (although with potentially sub-optimal performance).
For example, an individual baseline may be limited to a particular area of the network, and therefore
is only considered safe for that area, while other baselines may be safe for other areas.
Similarly, different \textit{RL agents} may follow different models or have different hyper-parameters.
In those circumstances, our approach benefits from the ability to compare multiple agents and baselines.
The modularity of the proposed solution enables customisation by modifying the shield logic module based on the use case and available RL agents and baselines.

\section{Related Work}
\label{sec:related_work}


Recently, many works have explored the use of RL methods as a solution for the RET optimisation problem \cite{guo2013spectral,balevi2019online,razavi2010fuzzy,fan2014self}. 
In \cite{guo2013spectral}, the authors maximise the UE's throughput fairness and energy efficiency of the network using an RL-based approach.
%
\cite{balevi2019online} addresses a similar problem while expanding the scope to a group of cells, each with its own antenna configuration, resulting in an extremely complex problem.
Both \cite{razavi2010fuzzy} and \cite{fan2014self} use Fuzzy Reinforcement Learning methods, utilising fuzzy control theory through the use of predefined rules in order to model the inherent uncertainty of the observed states of cellular networks and reduce the state dimensionality.
The main limitation with the majority of this work is that the solutions propose potentially unsafe actions that can have an adverse impact on the network.

Therefore, several methods have been proposed to achieve safety in the context of Safe Reinforcement Learning (SRL) \cite{garcia2015comprehensive}.
Examples of such safe strategies include the use of accumulated past knowledge \cite{ghavamzadeh2016safe}, making conservative action choices that prioritise avoiding the worst-case scenario, or requesting guidance from an external agent or a human operator when the current state is considered too risky \cite{torrey2012help, reddy2019learning}.
In \cite{alshiekh2018safe}, the definition of shielding is introduced where temporal logic, a model-based method, is proposed to filter out unsafe actions proposed by the RL agent.
In \cite{vairamuthu2020antenna}, the authors consider an SRL method based on an actor-critic model where a safety threshold $\varepsilon$ is progressively adjusted during the training, which only allows actions that are close to that of a safe baseline policy. The authors of \cite{Vannella21} tackle RET optimisation using an offline RL algorithm that modifies the exploration based on the information contained in a batch of data. Such method can be framed as a particular instance of our architecture with a single baseline policy and a single learning agent. 

%
%
%
In \cite{zhang2019towards}, the authors use a linear combination of suggested actions from the RL agent or the baseline with a parallel pioneer policy to prevent drastic changes in the control policy. 
Many of these works are tied to a specific RL algorithm, and none of them consider multiple baselines for increased safety in state spaces where an individual baseline might be inadequate. 
Our approach tackles these shortcomings in the current state of the art by proposing an SRL architecture that provides the flexibility to plug in various types of RL agents, baselines and decision logics towards better safety guarantees.


\section{Reinforcement Learning}
\label{sec:background}

Reinforcement Learning (RL) \cite{kaelbling1996reinforcement} is a long-established decision-making framework, in which an agent interacts with an environment by exploring its states and selecting actions to maximise the long-term return based on a reward signal.
%
%
An RL agent follows a policy $\pi$ choosing the action to be performed at each state.
The agent devises an optimal policy by dynamically optimising a value function that is estimated based on previous interactions of the RL agent.
At each time step, the agent can choose to either follow the recommended action from the estimated best action (exploitation), or choose a random action to gather further information about the environment and update its knowledge (exploration).

There are multiple approaches to find the optimal policy.
$Q$-learning is one of the most popular RL algorithms that finds the optimal policy by estimating the state-action value function ($Q$-function) $Q^{\pi(s,a)}$, representing the expected cumulative reward when being in state $s$, executing action $a$, and following policy $\pi$ afterwards.
%
%
%
%
When Artificial Neural Networks (ANNs) are used to parameterise the $Q$-function, the algorithm is known as Deep $Q$-Network (DQN) \cite{mnih2015human}. DQN uses optimisation methods such as Stochastic Gradient Descent (SGD) to minimise Mean Squared Error (MSE) between target values 
and the parameterised $Q$-function.
%
Another popular approach to find the optimal policy is the Actor-Critic (AC) algorithm. AC  uses ANN to parameterise two networks: the \textit{actor network}, taking as input the state and returning the action to be performed on the environment, and the \textit{critic network}, evaluating the action proposed by the actor by computing the value function. In this study we investigate DQN and AC agents for our proposed SRL architecture.


\section{A Safety Shield Architecture for Safe Reinforcement Learning}
\label{sec:architecture}

\begin{figure}[tbp]
\centering
    \vspace{0.04in}
    \includegraphics[width=.5\textwidth]{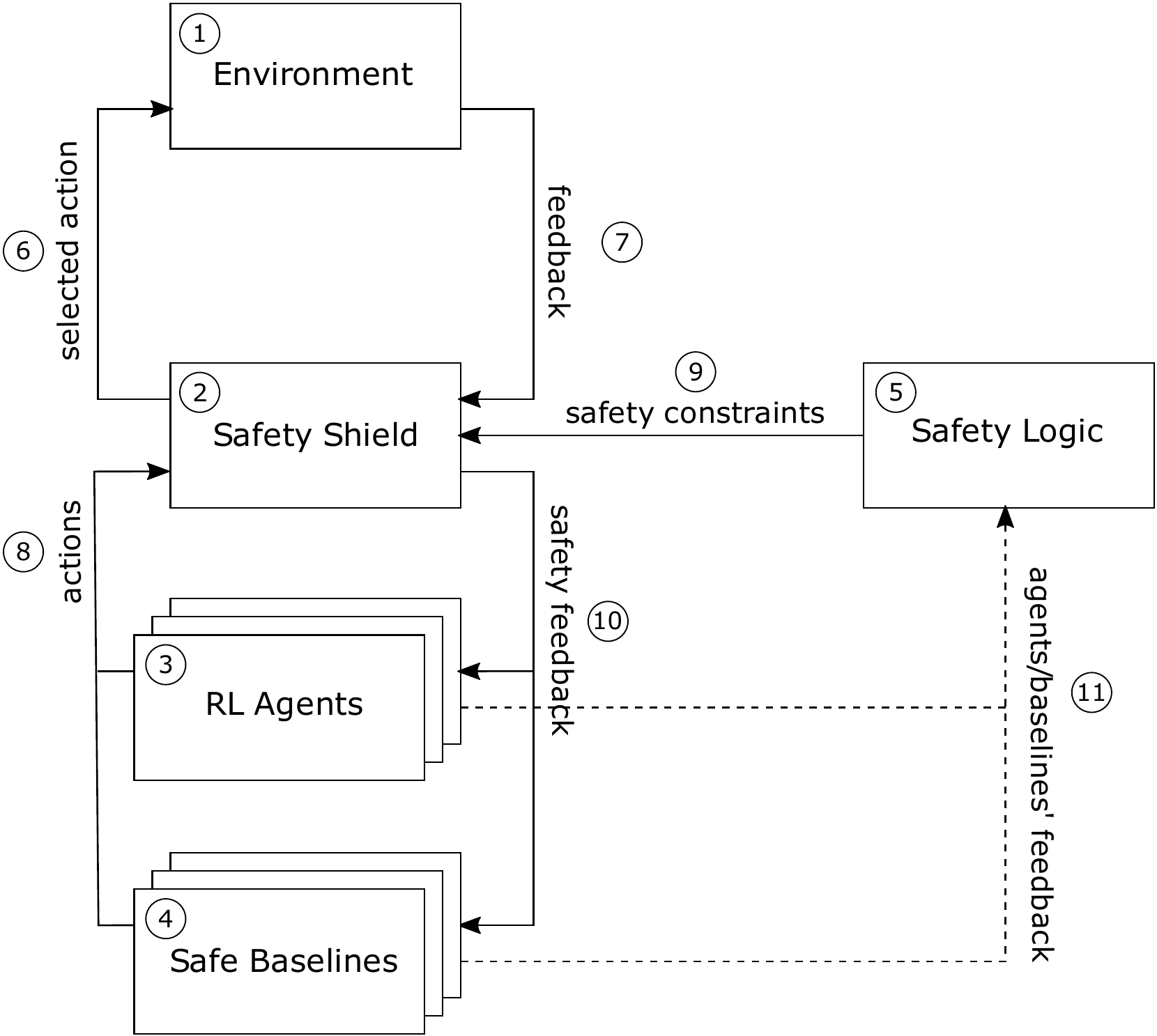}
\caption{SRL architecture with safety shield, logic and multiple baselines.}
\label{fig:architecture}
\end{figure}

Our approach is an SRL model that uses a modular shield architecture with one or more RL agents and baselines.
Figure \ref{fig:architecture} illustrates an overview of the proposed modular SRL structure.
It comprises a set of RL agents that are benchmarked against safe baselines through a safety shield that chooses the action to be performed on the environment.
The architecture components are the following (numbers refer to Figure \ref{fig:architecture}):

\begin{enumerate}[(1)]
    \item \textbf{Environment}: This component represents the environment that the agent is interacting with, modelled as a standard RL problem \cite{kaelbling1996reinforcement}.
    \item \textbf{Safety Shield}: This component is the mediator between the RL agents (3) and the environment (1).
    Contrary to the traditional RL model, where agents directly interact with the environment and receive a feedback, the safety shield acts as a proxy between agents and the environment to protect it from possibly unsafe actions proposed by the agents.
    The safety shield collects proposed actions by agents and baselines (8) and chooses a final safe action (6) to be performed on the real environment.
    The environment feedback (7) includes the action that is actually performed by the shield on the environment and is fed back to agents and baselines through the safety feedback (10).
    \item \textbf{RL Agents}: A set of RL agents indirectly interact with the environment by proposing actions (8) to the safety shield.
    The agents are continuously trained during the RL interaction to improve their future recommended actions.
    Using different learning methods enables parallel experimentation with multiple techniques and parameters, to determine the suitable safe action amongst them.
    \item \textbf{Safe Baselines}: Baselines also receive feedback about the current state of the environment to suggest future actions.
    Contrary to the RL agent actions, baseline actions are considered safe at least on a subset of the state-action space.
    Examples of safe baselines are rule-based algorithms that recommend actions satisfying safety rules designed by domain experts and supervised models trained on previous safe actions.
    The baselines are considered sub-optimal in terms of maximum performance but safe with regards to the minimum level of performance they provide at any given time.
    \item \textbf{Safety Logic}: The logic provides the safety constraints (9) to the safety shield, which are used to select the safe action from agents and baselines.
    The architecture supports various safety logic implementations, for example $(i)$ using only the proposed actions from agents and baselines, where the safety logic chooses the best action by some safety criteria,
    or $(ii)$ using agent and baseline internal variables such as convergence as additional information to gauge the safety of the proposed action.
    This is enabled through the feedback function. The shield logics investigated in this study are of type $(i)$.
\end{enumerate}

\section{Antenna Tilt Optimisation using SRL}
\label{sec:safe-rl-ret}


{The definition of RET optimisation problem follows previous works for \textit{Coverage-Capacity Optimization} (CCO) \cite{Buenostado17, Vannella20}, where the goal is to maximise network \textit{coverage} and \textit{capacity}, while minimising inter-cell \textit{interference}. We define network measurements for capacity and coverage KPIs for time $t$ and cell $c$ to be $\textsc{Cov}_{t,c}$ and $\textsc{Cap}_{t,c}$ respectively. Also, the effect of negative interference on neighbouring cells is modelled by the {quality} KPI $\textsc{Qual}_{t,c}$. Note that above metrics are considered risk KPIs, i.e. high values indicate high risk for insufficient coverage, capacity or quality in a given cell.}

\begin{table}[tbp]
    \centering
    \vspace{0.04in}
    \caption{Parameters for the simulator and the RL training.}
    \label{tab:parameters}
    \begin{tabular}{lr}
        \hline
        \textsc{Simulator parameter} & \textsc{Value} \\ 
        \hline
        Number of base stations & $7$ \\
        Number of cells & $21$ \\
        Number of user equipments   & $2000$ \\
        Frequency  & $2$ GHz \\
        Traffic volume & $20$ Mbps \\
        Antenna height & $32$ m \\
        Minimum downtilt angle & $1^{\circ}$\\
        Maximum downtilt angle & $16^{\circ}$\\
        \hline
        \textsc{DQN Hyper-parameter}       & \textsc{Value} \\
        \hline
        Discount factor & $0.0$ \\
        Learning rate & $0.001$ \\
        Batch size & $50$ \\
        Test episode length & $20$ \\
        Number of evaluation episodes & $25$ \\
        \hline
        \textsc{AC Hyper-parameter}       & \textsc{Value} \\
        \hline 
        Discount factor & $0.0$ \\
        Learning rate & $0.03$ \\
        Test episode length & $20$ \\
        Number of evaluation episodes & $25$ \\
        \hline 
    \end{tabular}
\end{table}

We now formulate the RET optimisation problem as a use case for the SRL architecture introduced in the previous section, with the following components:

\begin{itemize}
    \item \textit{Environment}: consists of a simulated mobile network environment in an urban area. The parameters of this environment are detailed in Table \ref{tab:parameters}. 
    Further, to avoid local optima, the cell downtilt values are reset to random values at the end of each episode.
  \item \textit{State space:} $\mathcal{S} \subseteq [0,1]^4$: comprises normalised values of \textit{downtilt} $\theta$, \textit{coverage} \textsc{Cov}, \textit{capacity} \textsc{Cap} and \textit{quality} \textsc{Qual} for time $t$ and cell $c$., i.e. $s_{t,c} = [\theta_{t,c},\textsc{Cov}_{t,c},\textsc{Cap}_{t,c}, \textsc{Qual}_{t,c}]$. 
  \item \textit{Action space:} $\mathcal{A} =\{-1,0,1\}$: consists of discrete changes of the current downtilt. At time $t$ and at cell $c$ the agent selects $a_{t,c}\in\mathcal{A}$.
  \item \textit{Reward:} $r_{t,c} = -\log(1 + \textsc{Cov}_{t,c}^2 +\textsc{Cap}_{t,c}^2 + \textsc{Qual}_{t,c}^2)$: consists of the squared log-sum of coverage, capacity, and quality risk KPIs.
  \item \textit{Safe baselines}: we considered two baselines for the experiments conducted for this study:
    \begin{itemize}
        \item \textit{Rule-based baseline:} this baseline is based on a legacy RET algorithm currently in place for antenna tilt optimisation that suggests antenna tilt actions based on $\textsc{Cov}_{t,c}$, $\textsc{Cap}_{t,c}$ and $\textsc{Qual}_{t,c}$.
        \item \textit{Model-based baseline:} this baseline is generated from a DQN-based offline learner trained on historical data of performed antenna tilts.
     \end{itemize}
  \item \textit{RL Agents:} In the experiments conducted for this study, we use an RL agent using the DQN or the AC algorithm.
  The training parameters of these agents 
  are 
  given in Table \ref{tab:parameters}. {The same policy is independently applied to all cells to suggest a specific action for each cell.}
\end{itemize}




In this setting, we study the performance of two shield logics, namely State Predictor Shield Logic and $k$-Shield Logic, which are described below.

\subsection{State Predictor Shield Logic}
\label{sec:state-predictor}

The State Predictor Shield Logic uses a pre-trained supervised learning model to assess the value of the actions proposed by the different RL agents and baselines.
    We train a multi-target regressor on historical data from the environment, where the input for the model is defined as a given environment state plus a tilt change action $(\textsc{Cov}_{t,c}, \textsc{Cap}_{t,c}, \textsc{Qual}_{t,c}, a_{t,c})$, and the output is defined as the environment state after applying the action $(\textsc{Cov}_{t+1,c}, \textsc{Cap}_{t+1,c}, \textsc{Qual}_{t+1,c})$.
The regressor in our experiments is a fully connected multi-layer perceptron.
We also experimented with extended state-space, including additional KPIs, but that did not noticeably impact the prediction accuracy of the model.

With a sufficiently large sample size, a supervised model can be trained that captures the impact of various antenna angles on the measured network KPIs with reasonable accuracy.
This model is of course subject to the limitations of supervised learning that RL is trying to solve, such as decreased accuracy on unseen data, because supervised learning is susceptible to wrong/incomplete data and cannot adapt from interaction with the environment.
For the results reported in this work, we train the supervised State Predictor model on data that we synthesised from the simulation environment by randomising environment parameters to create a varied set of data points with different antenna tilt angles and states.
This way, the data for the supervised model follows a similar distribution to the data the RL is operating on. 
However, we also experimented with a supervised model trained on historical data from a real network deployment, and we found that it performed on a similar level compared to the one trained on synthetic data.
This underlines the viability of using supervised learning models, even from a different deployment environment, for the purpose of increasing safety of the RL approach.

The State Predictor Shield Logic uses this pre-trained model to compare the actions proposed by agents and baselines by querying the model with each action in addition to the current environment state.
From the predicted next state, the shield logic can calculate various metrics to score the proposed actions, such as a reward or a safety constraint.
In our experimental setup, the safety logic compares proposed actions by using the state KPI directly.
It will choose the action that achieves the best predicted state KPI and forwards it to the shield instance for execution on the environment.

\subsection{$k$-Shield Logic}
\label{sec:K-Shield}
The logic in this scenario is inspired by the approach proposed in \cite{zhang2019towards} and is generalised to accommodate multiple
safe baselines.
It chooses an action by selecting either the RL agent's action or one of the baselines' action with a given probability, gradually switching control over from the baselines to the RL agent.
Formally, the control policy $\pi_\mathcal{C}$ is chosen between the RL policy $\pi_L$ and a set of $B$ safe baseline policies $\{\pi_s^1, \dots, \pi_s^B\}$ as follows:



\begin{equation}
    \pi_\mathcal{C} = p \pi_s^i + (1-p) \pi_L.
\end{equation}

When selecting between the safe baselines or the RL policy, we sample $p\sim \mathcal{B}(k)$, i.e. a Bernoulli random variable with parameter $k \in [0,1]$.
The parameter $k$ determines the probability of choosing (one of) the safe baselines over the RL policy.
The probability $k$ is initially close to $1$ to ensure that the control policy is dominated by safe baselines when the RL policy has not accumulated enough experience to be able to execute safe actions.
Subsequently, its value is reduced by a diminishing factor $d\in (0,1)$ every time the reward during previous episodes is improved compared to the past $w$ episodes. 
Each episode is indexed as $e = 1,\dots, E$. The update is executed each $w$ episodes as:
\begin{equation}
  k_{e+1} =
    \begin{cases}
      \max\{0,k_e - d\} & \hat{R}_{[e-w,e]} \geq \hat{R}_{[e-2w,e-w-1]} \\
      k_e & \text{otherwise}
    \end{cases}
\end{equation}
where $\hat{R}_{[l,m]}$, denotes the average reward calculated from episode $l$ to  episode $m$.

The categorical random variable $i \sim \textsc{Cat}(b)$, with parameter $b \in [0,1]^B: \sum_{i=1}^B b_i = 1$, controls the importance of each baseline. If $p=1$, we select the $i$-th baseline with probability $b_i$.
%
%
The baselines probabilities $b_i$ can be:
\begin{itemize}
    \item preset values based on the performance of safe baselines, and defined by domain experts or use-case stakeholders (e.g., network operators in RET optimisation), or 
    \item state-action dependent, allowing higher weights for policies that outperform others in certain sub-spaces of the problem (e.g., if a safe baseline has higher performance on a sub-region of a cellular network).     
\end{itemize}

 
In the experiments conducted for this study, we investigate scenarios where an RL agent competes with a single baseline and with multiple baselines for different values of $b$.
The results of our experimentation 
are detailed in the next section.


\section{Evaluation Results}
\label{sec:experiment_results}


The results discussed in this section present the outcomes of our shielded SRL experiments using the modular shield architecture described in Section \ref{sec:architecture}.
We run each experiment with $6$ random seeds, and
each plot shows the average of all random seeds (continuous and dashed lines), as well as the minimum and maximum values that were reached over all seeds (shaded areas).
We apply a running average window to smooth out the curves for better visual comprehension.
Finally, we focus on the early phase of RL training, in which exploration is prevalent and therefore most unsafe behaviour of an unrestricted RL agent is to be expected.
The long-term performance advantages of RL over traditional approaches have already been shown in the literature, e.g. \cite{guo2013spectral,razavi2010fuzzy,fan2014self}.

The following elements will be shown in the plots:
\begin{enumerate}
    \item Unrestricted RL agents with full environment access,
    \item Baselines that are considered sub-optimal but safe,
    \item SRL agents that are governed by the safety shield using a shield logic.
\end{enumerate}
The plots focus on reward (higher is better), and coverage and quality risk KPIs (lower is better). 
The capacity KPI is omitted as it was not impacted by antenna tilts across all experiments.

\subsection{Comparison of unrestricted RL agent with baseline and SRL agent}
In this section we compare the performance of unrestricted RL agents with baselines and with the SRL agents using the two shield logics described in sections \ref{sec:state-predictor} and \ref{sec:K-Shield}.
%
 
\begin{figure}[tbp]
    \centering
    \begin{subfigure}[t]{0.49\columnwidth}
        \includegraphics[width=\textwidth]{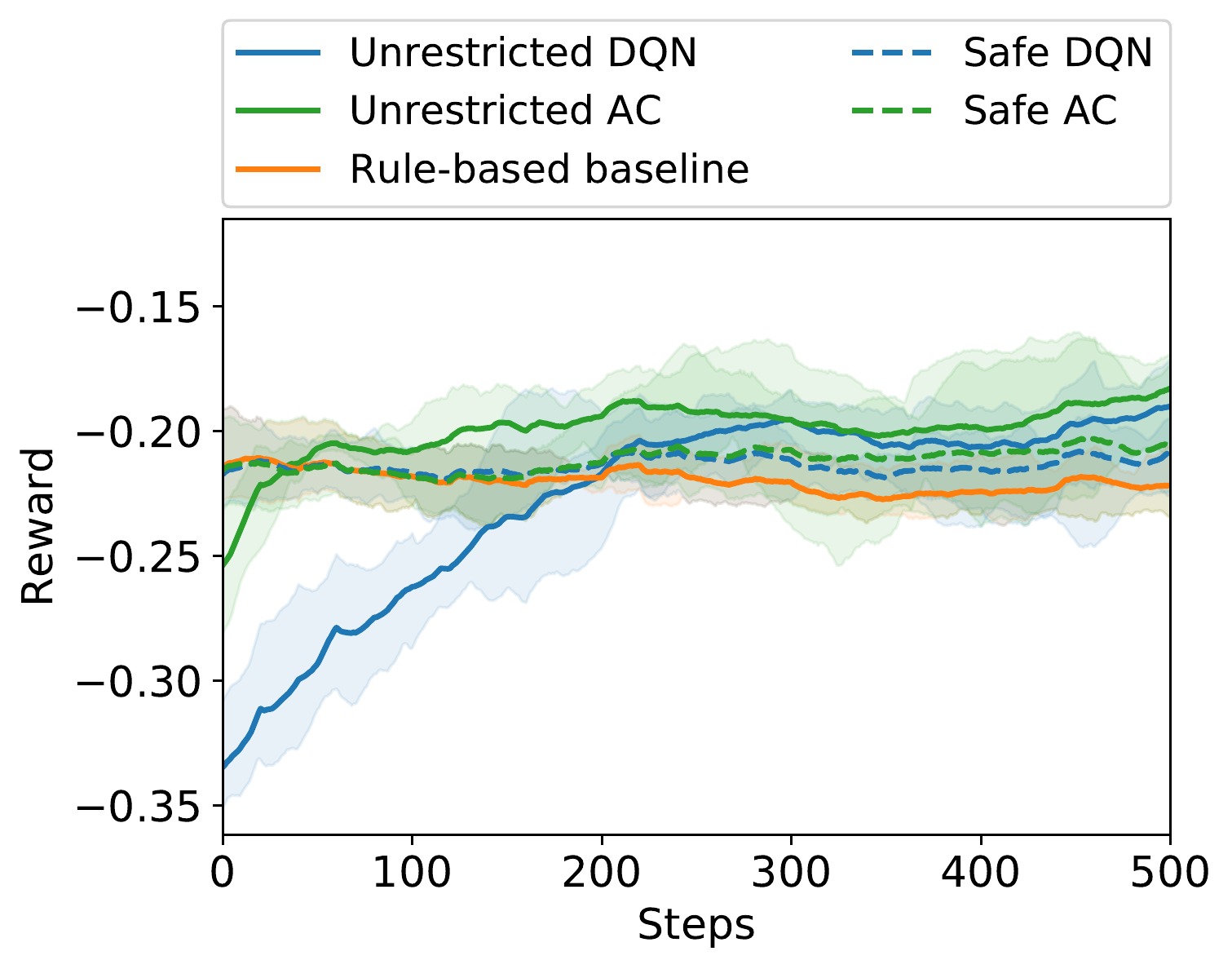}
    \end{subfigure}
    \begin{subfigure}[t]{0.49\columnwidth}
        \includegraphics[width=\textwidth]{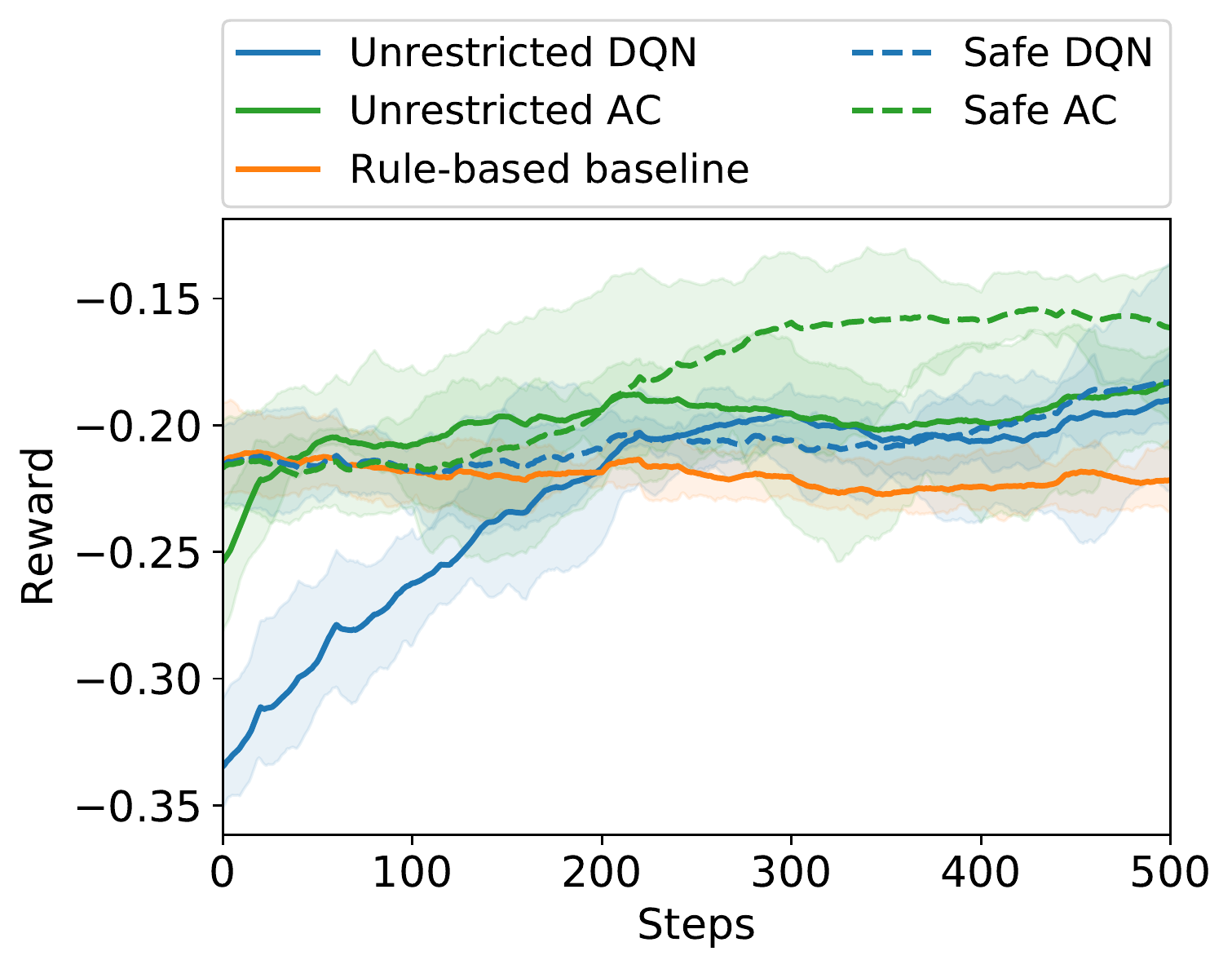}
    \end{subfigure}
    \begin{subfigure}[t]{0.49\columnwidth}
        \includegraphics[width=\textwidth]{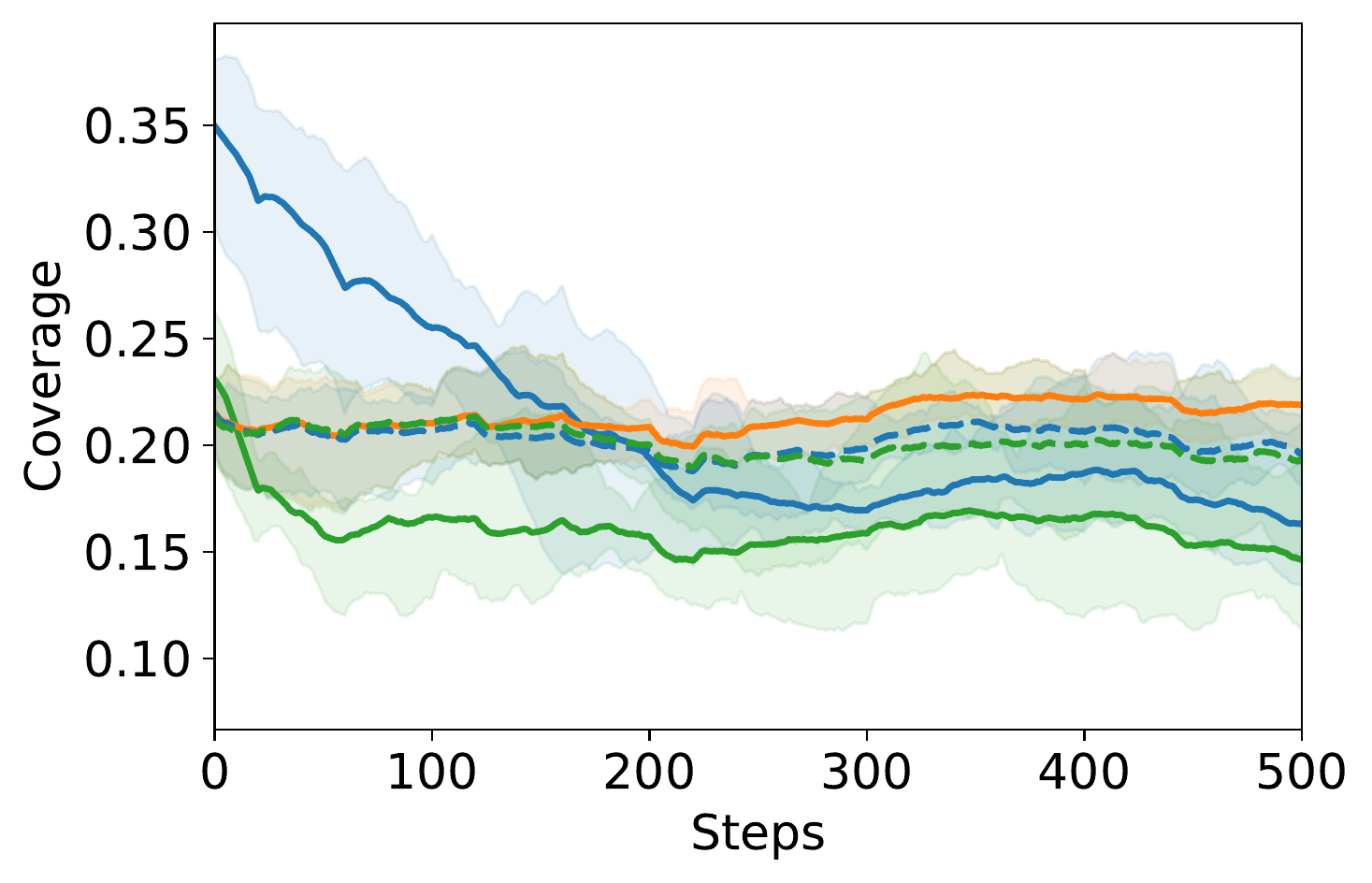}
    \end{subfigure}
    \begin{subfigure}[t]{0.49\columnwidth}
        \includegraphics[width=\textwidth]{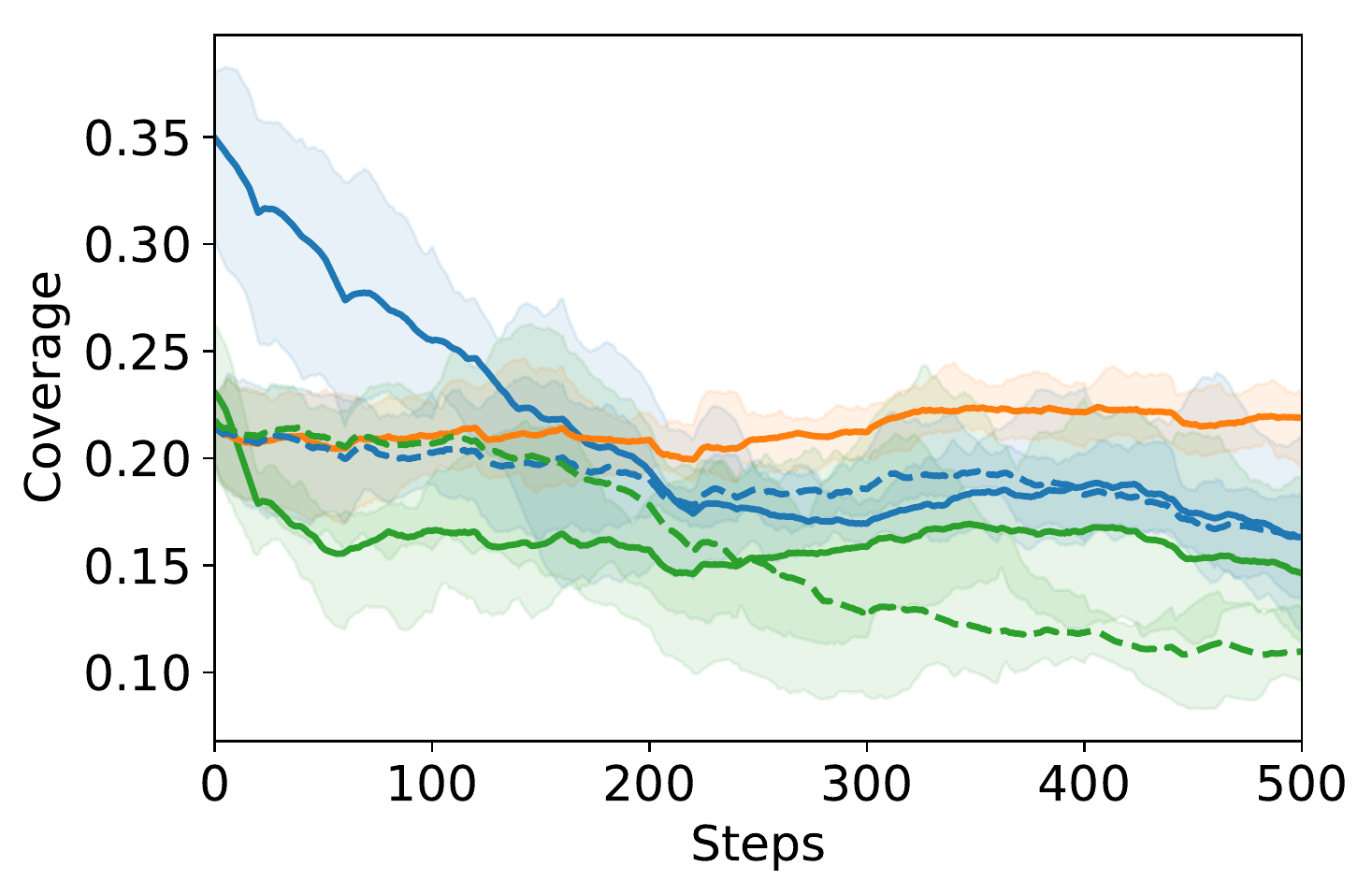}
    \end{subfigure}
    \begin{subfigure}[t]{0.49\columnwidth}
        \includegraphics[width=\textwidth]{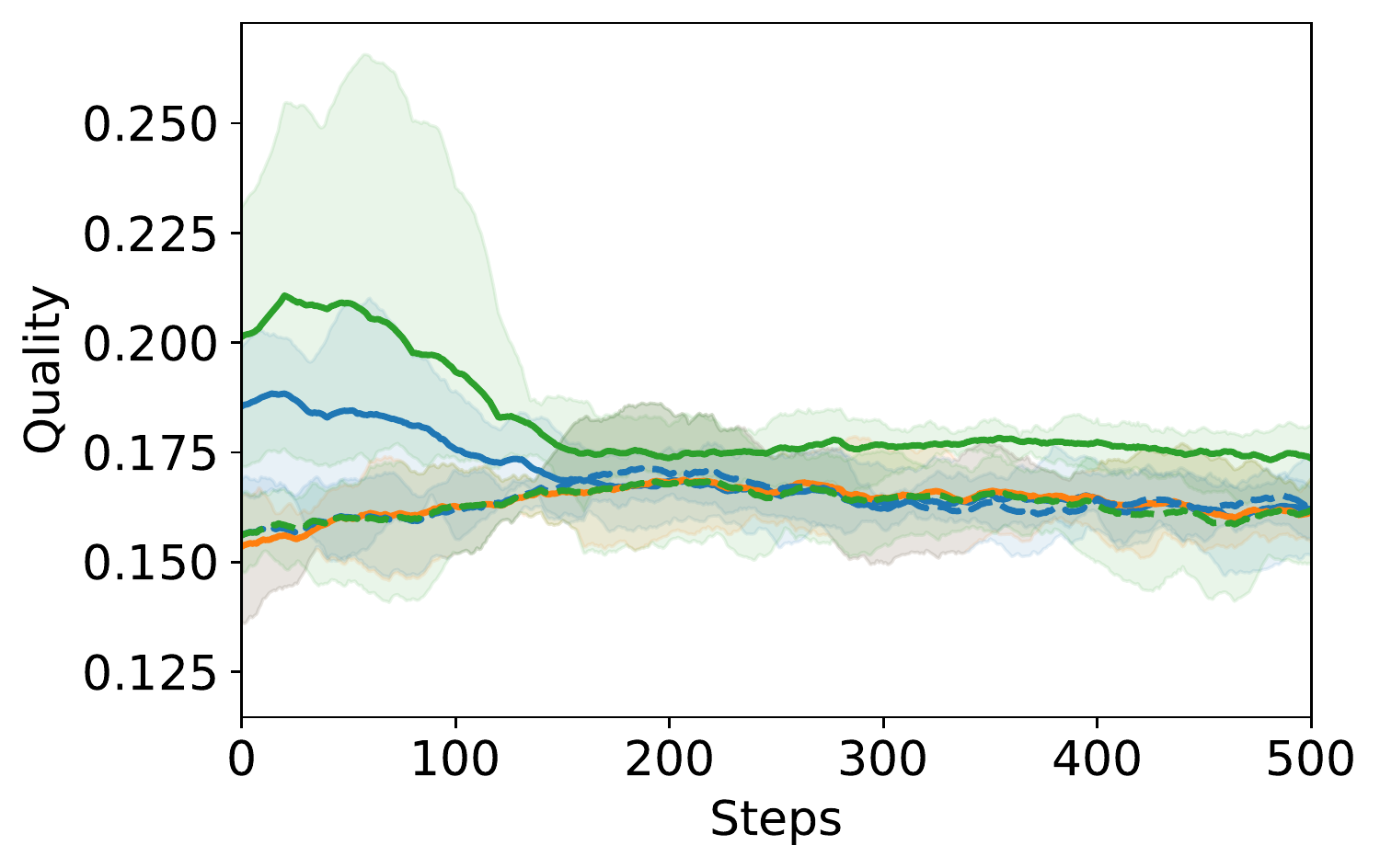}
        \caption{State Predictor Shield Logic.}
        \label{fig:state_predictor_results}    
    \end{subfigure}
    \begin{subfigure}[t]{0.49\columnwidth}
        \includegraphics[width=\textwidth]{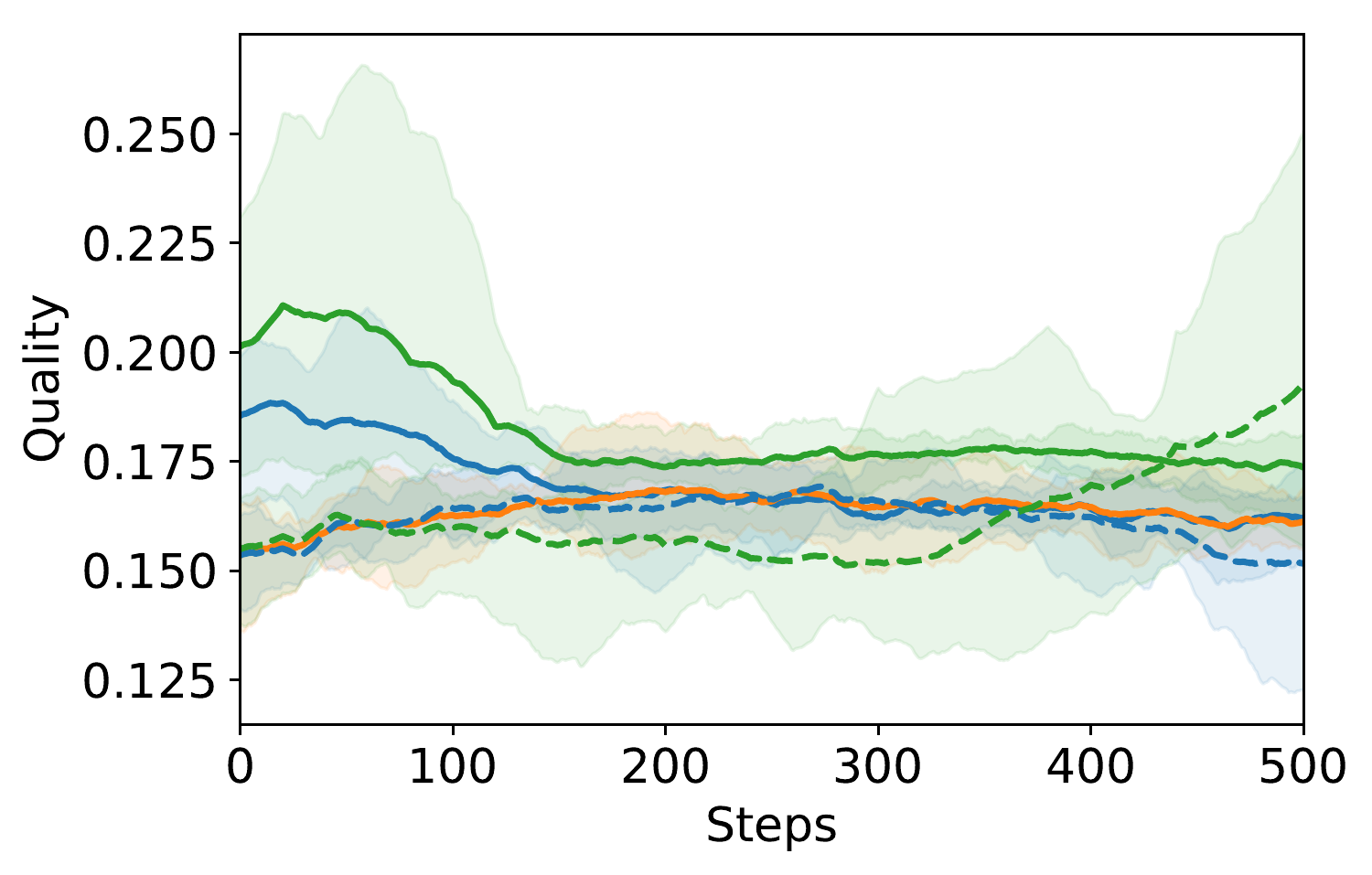}
        \caption{$k$-Shield Logic.}
        \label{fig:k_shield_results}    
    \end{subfigure}
    \caption{Comparison of reward (higher is better), coverage and quality risk KPIs (lower is better) for DQN and AC using unrestricted agents, baseline only and safe agents governed by the shield logic. Shaded areas represent min/max over all runs.}
    \label{fig:comparison_results}
\end{figure}

The State Predictor Shield Logic uses a pre-trained machine learning model to estimate the impact of the proposed tilt change action on the network as described in Section \ref{sec:state-predictor}.
Figure \ref{fig:state_predictor_results} (top) shows the immediate reward that is returned by the environment.
We observe that the unrestricted DQN agent (solid blue) starts off with a sub-optimal reward, which is significantly worse than the performance of the safe baseline.
When the safety shield is applied, the safe DQN agent (dashed blue) closely mimics the baseline behaviour and slightly outperforms the baseline early on.
This is the desired effect that the safety shield with the State Predictor Shield Logic was designed to accomplish.

Figure \ref{fig:state_predictor_results} (middle and bottom) focuses on coverage and quality, which are two important performance indicators.
In our experiments, coverage and quality are risk-based KPIs, which means that lower values indicate better network conditions. 
In both cases, it is apparent that the unrestricted DQN agent starts off at a much higher risk level before it learns the environment and its performance is improved, although the effect is more pronounced on the coverage KPI.
The safe baseline (orange) is at a safe level from the start with little fluctuation, where the safe level of performance is set by network operator experts.
The State Predictor Shield Logic causes the DQN agent to follow the baseline performance more closely from the start, thus creating a safe agent interaction with the environment. 
In the case of coverage, the safe agent slightly outperforms the baseline on average early on. 



Before comparing the performance of the $k$-Shield Logic, we first tune the diminishing factor $d$ and the average reward window size $w$ as described in Section \ref{sec:K-Shield}. We compared the performance of the safe DQN agent using $k$-Shield Logic for $d \in \{ 0.1, 0.2 \}$ and $w \in \{2,5\}$. While the results show very small difference,
a smaller $d$ indicates a more gradual transfer to RL agent's policy, allowing to rely on safe baselines for a longer period of the training. 
Also, smaller $w$ indicates more frequent comparison between average rewards over previous episodes, which allows quicker adaptation to current rewards, which in turn improves the performance of the agent governed by the $k$-Shield Logic.


Following this observation, $d=0.1$ and $w=2$ are selected as optimal parameter values that will be used for the $k$-Shield Logic experiments. Figure \ref{fig:k_shield_results} (top) shows the reward that is returned by the environment over time. As it can be observed, the safe DQN agent with $k$-Shield Logic (dashed blue) follows the safe baseline policy during the period in which the performance of unrestricted DQN (blue) is inferior to that of the the safe baseline. It then gradually switches towards unrestricted DQN agent as the model is further trained to achieve greater rewards than the safe baseline.
Similar to the State Predictor Shield Logic, for the coverage and quality risk KPIs (Figure \ref{fig:k_shield_results} middle and bottom), the $k$-Shield Logic prevents the DQN agent from performing high-risk actions by following the safe baseline policy at the start of the training phase, where the DQN agent performs mostly exploratory actions, and subsequently switching to the DQN actions when it has gathered enough experience to execute a safe action. This behaviour is more visible for coverage.




We also compare the unrestricted and safe AC agent to the unrestricted and safe DQN agent for both the State Predictor Shield Logic (Figure \ref{fig:state_predictor_results}, green) and the $k$-Shield Logic (Figure \ref{fig:k_shield_results}, green).
We observe that the trend for reward, coverage and quality KPIs when using a DQN agent is similar to when the safety logic is used to restrict actions of the AC agent, although the range of values for the two agent types differs. 
Both agent types converge during the experiments in terms of loss, although the shield tends to delay the convergence for the SRL agents slightly. 
We omitted the loss plots for brevity.

\subsection{Comparison of multiple baselines}

\begin{figure}[tbp]
    \centering
    \begin{subfigure}[t]{0.49\columnwidth}
        \includegraphics[width=\textwidth]{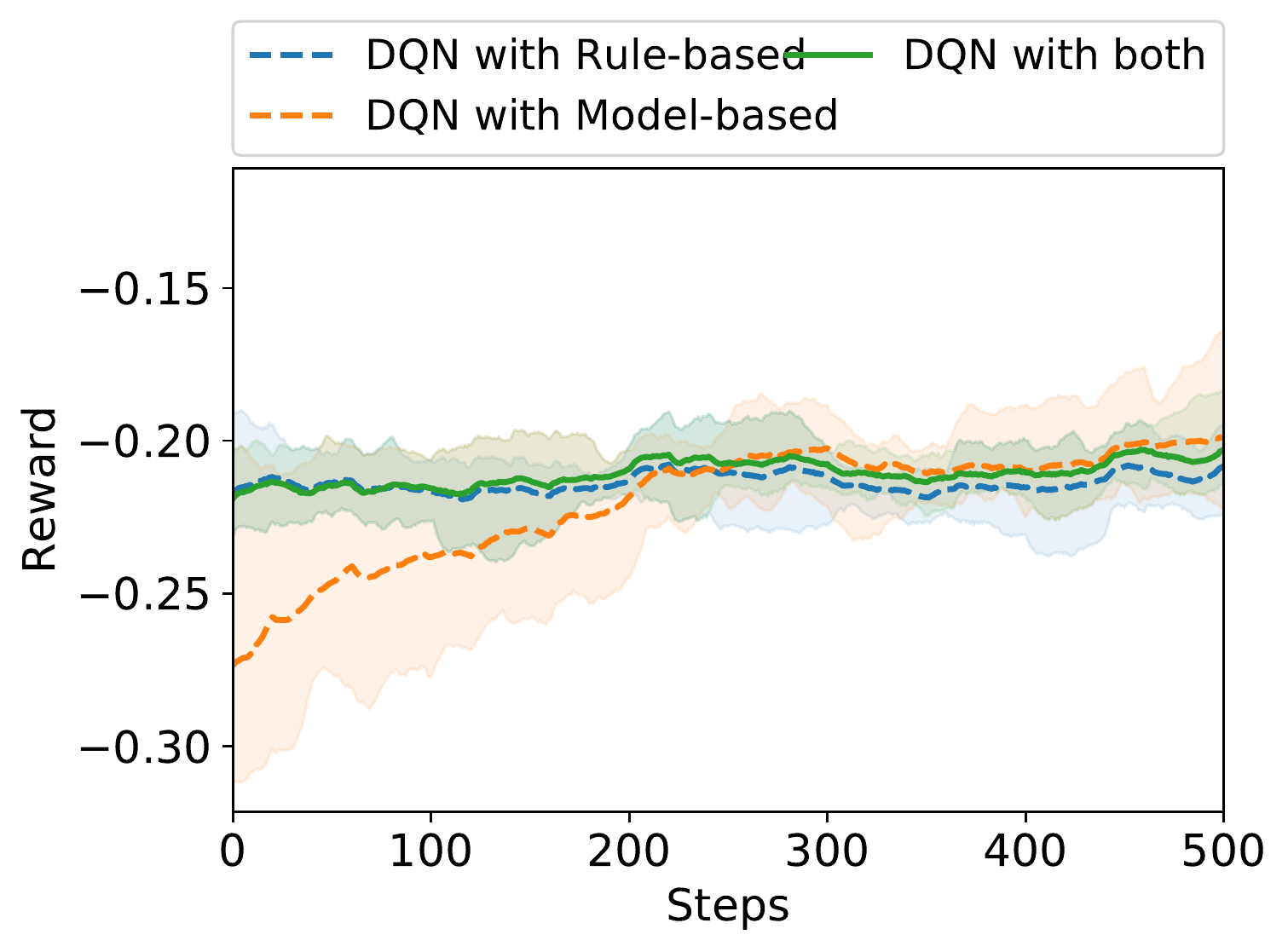}
    \end{subfigure}
    \begin{subfigure}[t]{0.49\columnwidth}
        \includegraphics[width=\textwidth]{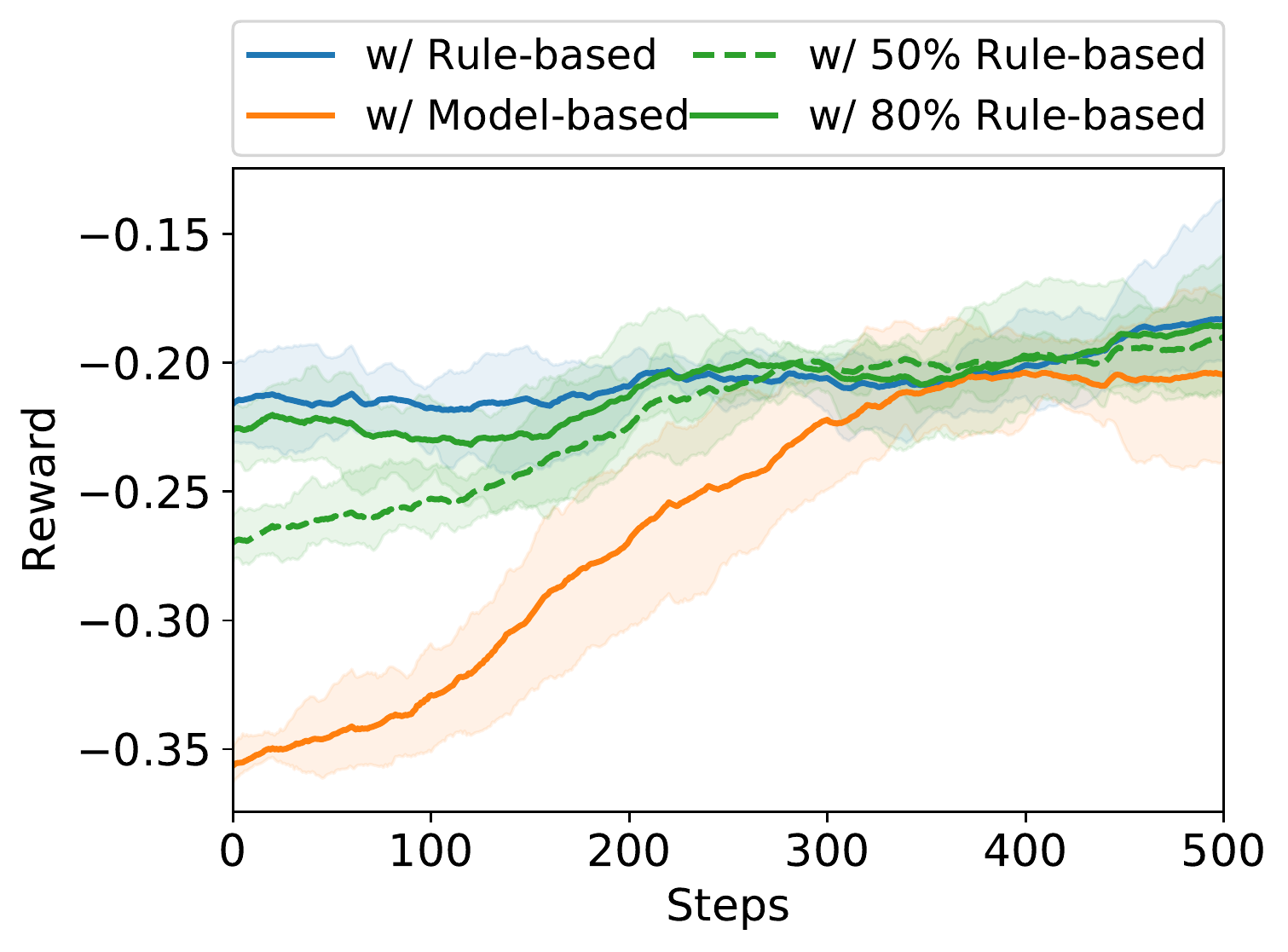}
    \end{subfigure}
    \begin{subfigure}[t]{0.49\columnwidth}
        \includegraphics[width=\textwidth]{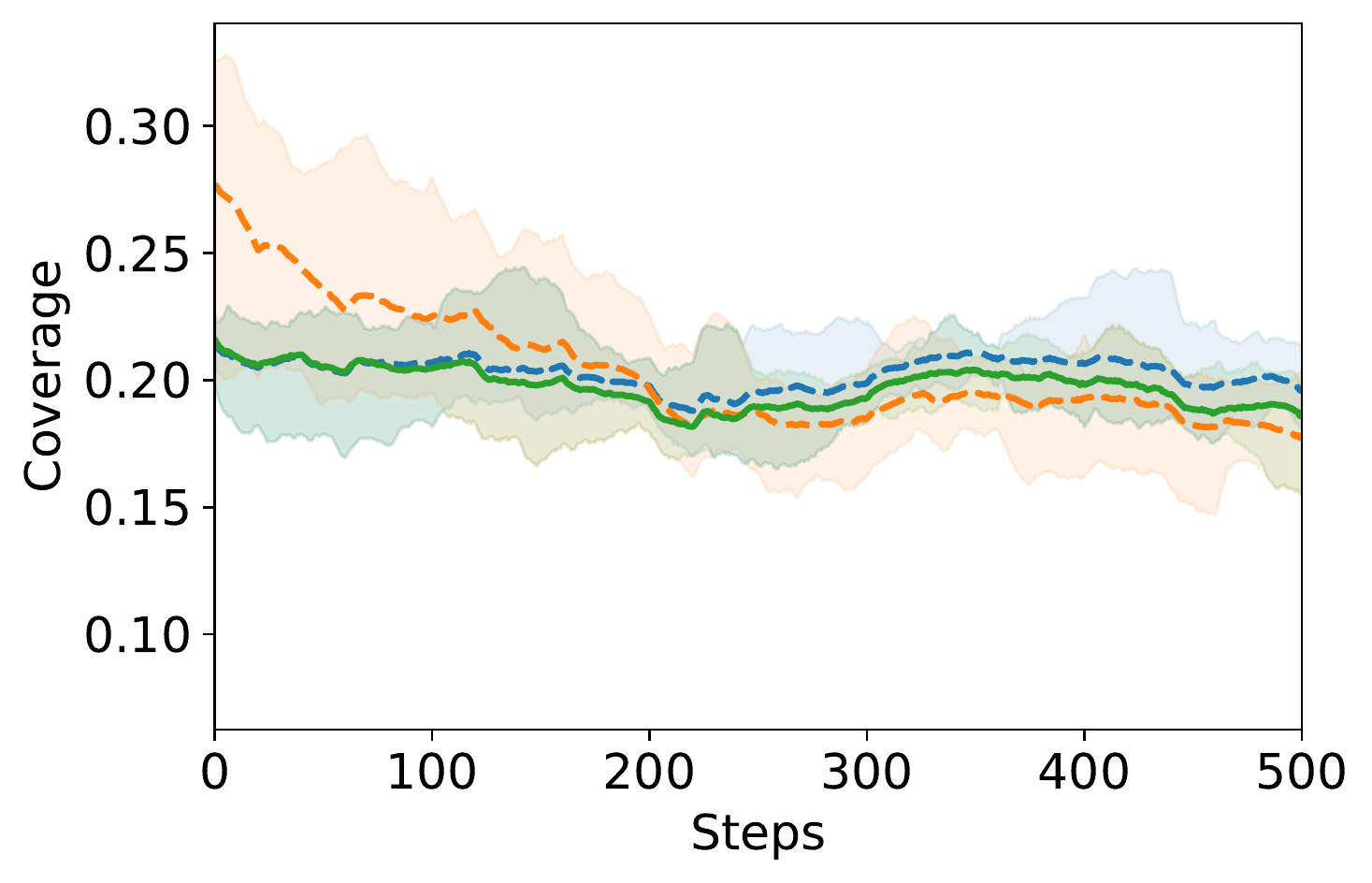}
    \end{subfigure}
    \begin{subfigure}[t]{0.49\columnwidth}
        \includegraphics[width=\textwidth]{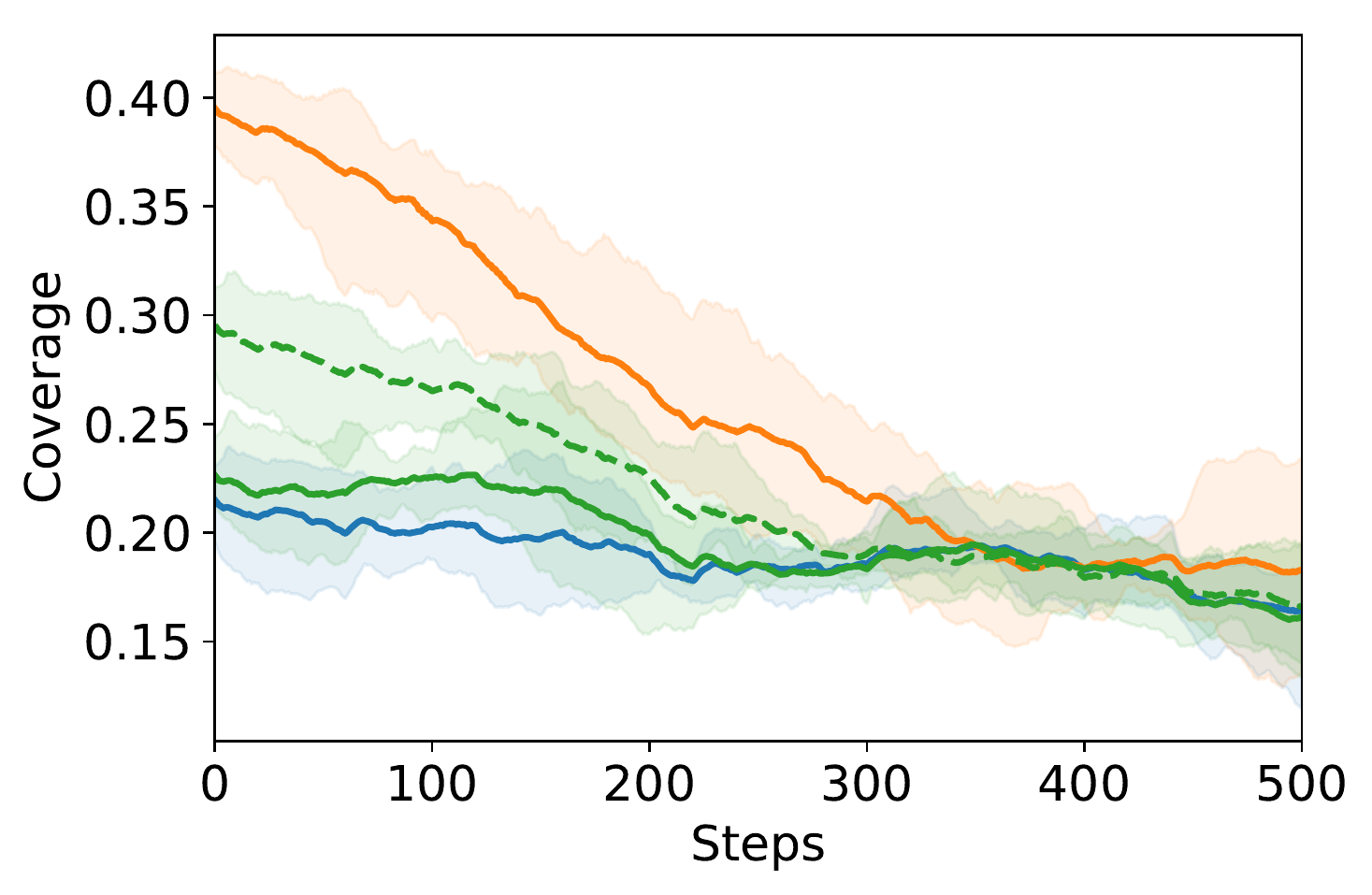}
    \end{subfigure}
    \begin{subfigure}[t]{0.49\columnwidth}
        \includegraphics[width=\textwidth]{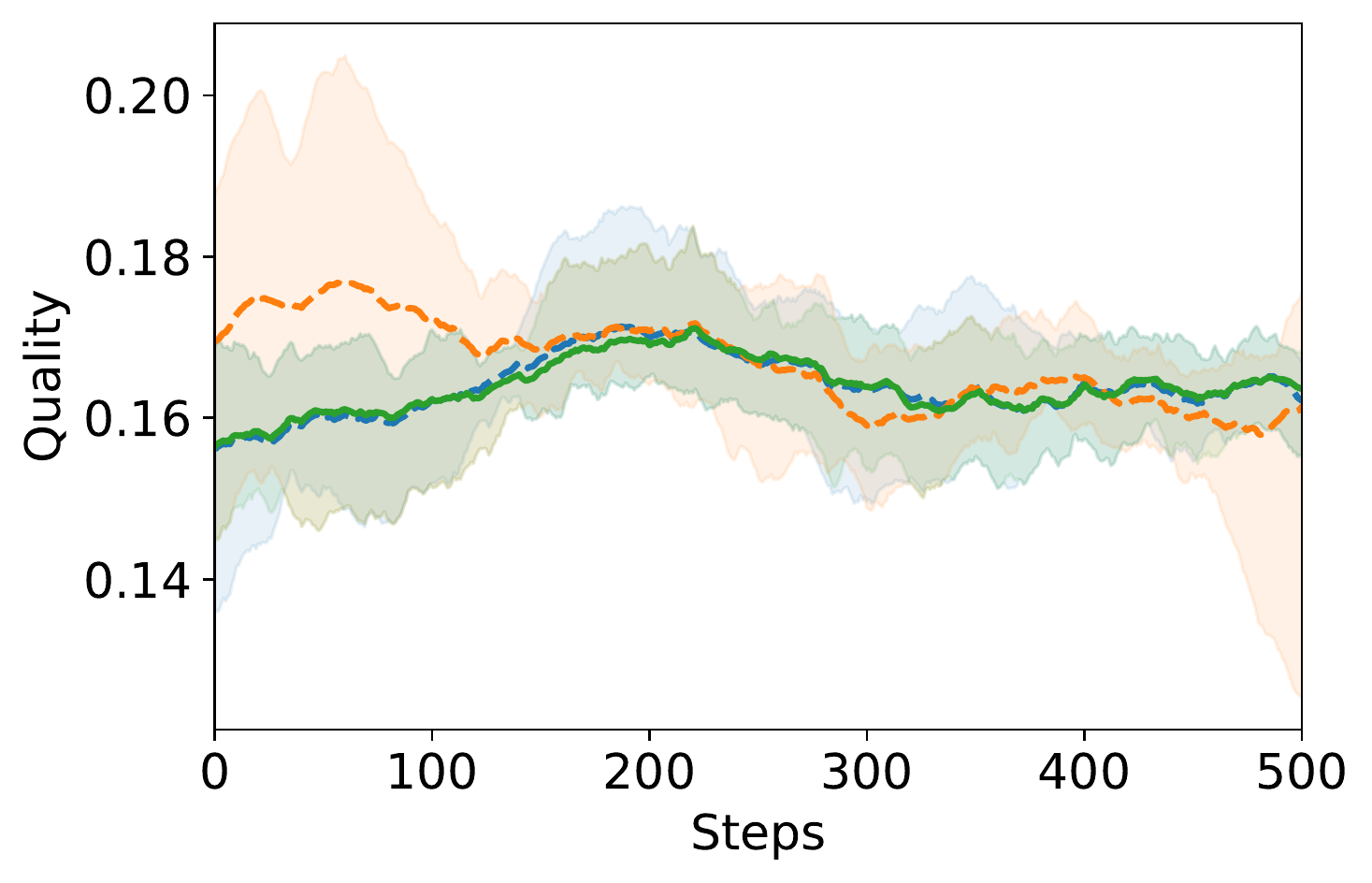}  
        \caption{State Predictor Shield Logic.}
        \label{fig:two_baselines_state_predictor_results}    
    \end{subfigure}
    \begin{subfigure}[t]{0.49\columnwidth}
        \includegraphics[width=\textwidth]{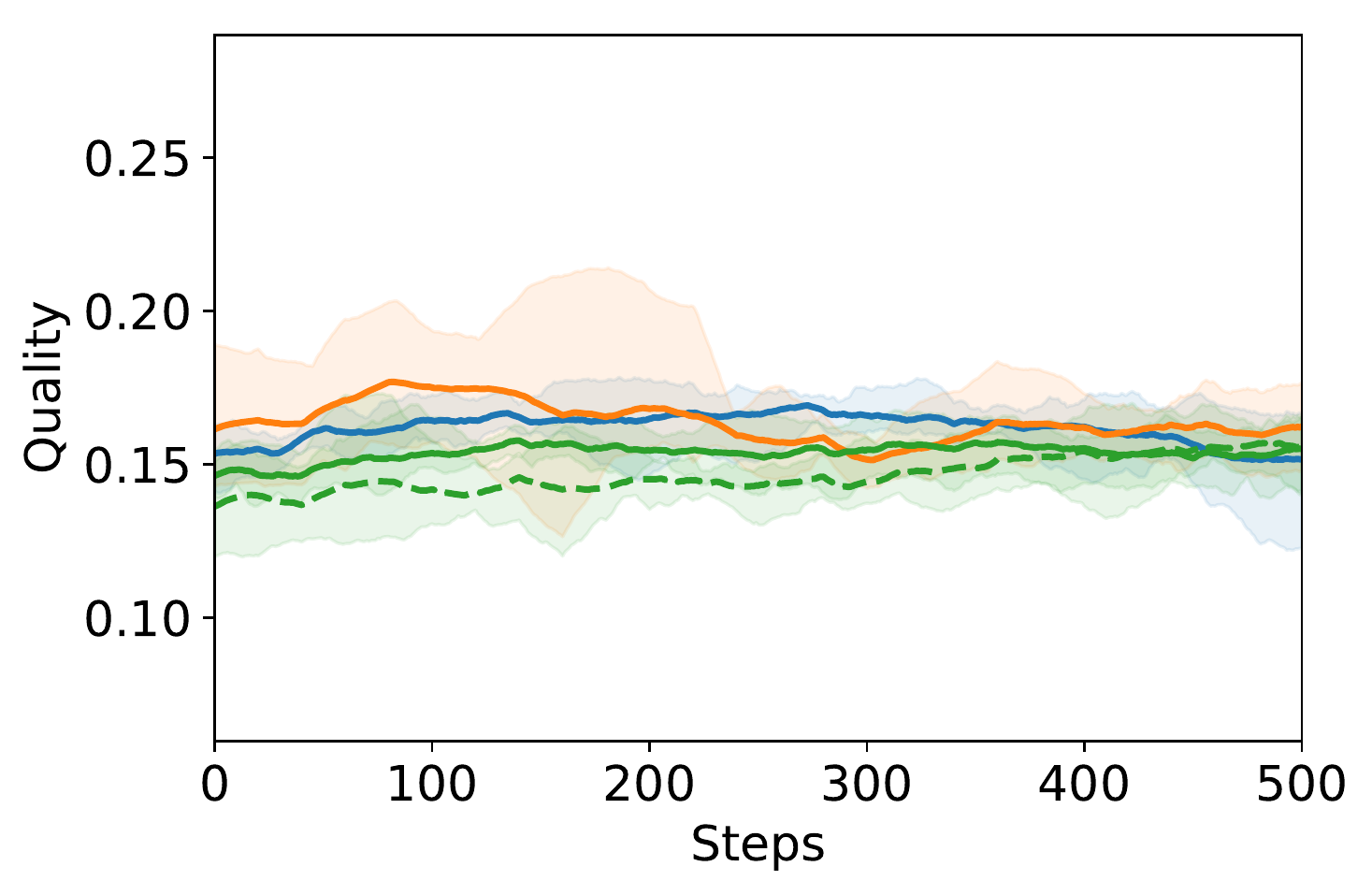}
        \caption{$k$-Shield Logic.}
        \label{fig:two_baselines_kshield_results}    
    \end{subfigure}
\caption{Comparison of reward (higher is better), coverage and quality risk KPIs (lower is better) for shield-logic governed DQN using rule-based, model-based and both baselines. Shaded areas represent min/max over all runs.}
\label{fig:two_baselines_results}
\end{figure}

In this section, we compare different baselines in a scenario where the shield logic has access to multiple baselines and can choose between them according to its strategy.
The two baselines are Rule-based and Model-based, as described in Section \ref{sec:safe-rl-ret}.
To keep the plots readable, these experiments focus on the DQN agent only.

Figure \ref{fig:two_baselines_state_predictor_results} shows a comparison of the received rewards and the achieved risk KPIs when training a DQN learner on a shielded environment where the State Predictor Shield Logic chooses between different baselines.
In the network environment considered for this study, the model-based baseline (dashed orange) performs worse than the alternative baseline (dashed blue).
However, when providing the State Predictor Shield Logic with both baselines, it is able to select the best proposed action based on the predicted state (green).


%
%
Figure \ref{fig:two_baselines_kshield_results} compares the reward, and coverage and quality risk KPIs of a $k$-Shield Logic restricted DQN agent using both Rule-based and Model-based baselines for different importance weights. 
It can be seen that in this scenario, where the Rule-based baseline is outperforming the Model-based baseline, higher weights on the Rule-based baseline result in faster convergence to better reward (top) and coverage (middle). This is less apparent for quality (bottom) as the difference between the performance for different scenarios is very small and within the margin of error.





\section{Conclusions}
\label{sec:conclusions}

In this paper, we propose a modular architecture for Safe Reinforcement Learning (SRL) that enables safe interaction with the environment by benchmarking recommendations of RL agents to multiple safe baselines.
The architecture allows experimentation with multiple RL agents, using different models and hyper-parameters, and leverages multiple baselines that may be safe in subsets of the state-action space, to create a policy that is safe across the whole state-action space and outperforms individual unrestricted RL agents and baselines.
As a concrete use case for our architecture, we study RET optimisation, which is evaluated using a simulated environment that is configured to recreate a real-world urban network.

The safety and performance of the approach using multiple shield logics is compared with an existing rule-based SON algorithm and an offline model that is trained on historical data.
The results show that, compared to safe baselines, unrestricted RL agents explore unsafe regions of the state-action space, especially at the start of the RL learning process.
The safety shield allows RL agents to start learning in safe states (performing similar to safe baselines) with the help of the shield logics, and eventually outperform safe baselines in most cases, especially for the received reward and the coverage KPI.
%
%
%
A limitation to this approach is the availability of safe baselines, without which the safety shield would not be able to decide whether a proposed action is safe enough for the environment.


\bibliographystyle{IEEEtran}
\bibliography{main}

\end{document}